\documentclass[a4paper, 10pt, conference]{ieeeconf} 
\IEEEoverridecommandlockouts
\usepackage{cite}
\usepackage{amsmath,amssymb,amsfonts}
\usepackage{algorithmic}
\usepackage{graphicx}
\usepackage{textcomp}
\usepackage{booktabs}
\usepackage{float}
\usepackage{hyperref}
\usepackage[ruled,vlined]{algorithm2e}
\usepackage{algorithm2e}
\usepackage[table]{xcolor}
\title{\LARGE \bf VerifyLLM: LLM-Based Pre-Execution Task Plan Verification for Robots}

\author{Danil~S.~Grigorev$^{1,2}$, Alexey~K.~Kovalev$^{3,1}$, and Aleksandr~I.~Panov$^{3,1}$
\thanks{$^{1}$MIPT, Dolgoprudny, 141701, Russia $^{2}$Pyatigorsk State University, Pyatigorsk, Stavropol Krai, 357532, Russia $^{3}$AIRI, Moscow, 121170, Russia {\tt\small \{kovalev,panov\}@airi.net}}%
}

\begin{document}
\maketitle

\begin{abstract}
In the field of robotics, researchers face a critical challenge in ensuring reliable and efficient task planning. Verifying high-level task plans before execution significantly reduces errors and enhance the overall performance of these systems. In this paper, we propose an architecture for automatically verifying high-level task plans before their execution in simulator or real-world environments. Leveraging Large Language Models (LLMs), our approach consists of two key steps: first, the conversion of natural language instructions into Linear Temporal Logic (LTL), followed by a comprehensive analysis of action sequences. The module uses the reasoning capabilities of the LLM to evaluate logical coherence and identify potential gaps in the plan. Rigorous testing on datasets of varying complexity demonstrates the broad applicability of the module to household tasks. We contribute to improving the reliability and efficiency of task planning and addresses the critical need for robust pre-execution verification in autonomous systems. The code is available at \url{https://verifyllm.github.io}.
\end{abstract}

\section{INTRODUCTION}
Verifying robot action plans before execution remains a challenging task in robotics~\cite{taioli2024mind, carta2023grounding}. Modern planning systems generate action sequences that appear correct at first glance but contain hidden errors that only become evident during execution. For example, a robot might attempt to pour water into an upside-down glass or a closed container. These errors do not stem from the robot's physical limitations, but rather from action plans that fail to incorporate common sense and fundamental physical constraints that humans naturally take into account.

Traditional planning systems based on PDDL~\cite{fox2003pddl2} struggle with tasks requiring common sense reasoning, often overlooking essential preconditions (e.g., checking whether a container is empty before filling it~\cite{sharma2022correcting}) and fail to consider action consequences(e.g., checking that the fridge is closed after interaction with it). Recent approaches have evolved from classical planning methods to learning-based techniques~\cite{ren2024extended}, which better handle uncertainty and complex environments. Large language model (LLM)-based frameworks have shown promising results for robotic task planning~\cite{huang2022language,kovalev2022application,sarkisyan2023evaluation,ni2024grid,onishchenko2025lookplangraph}. However, most studies focus primarily on generating plans rather than verifying them~\cite{huang2022language,ahn2022can,sarkisyan2023evaluation}, or provide verification mechanisms with limited scope due to their reliance on simple templates~\cite{grigorev2024common}.
\begin{figure}[t]
\centering
\includegraphics[width=\columnwidth]{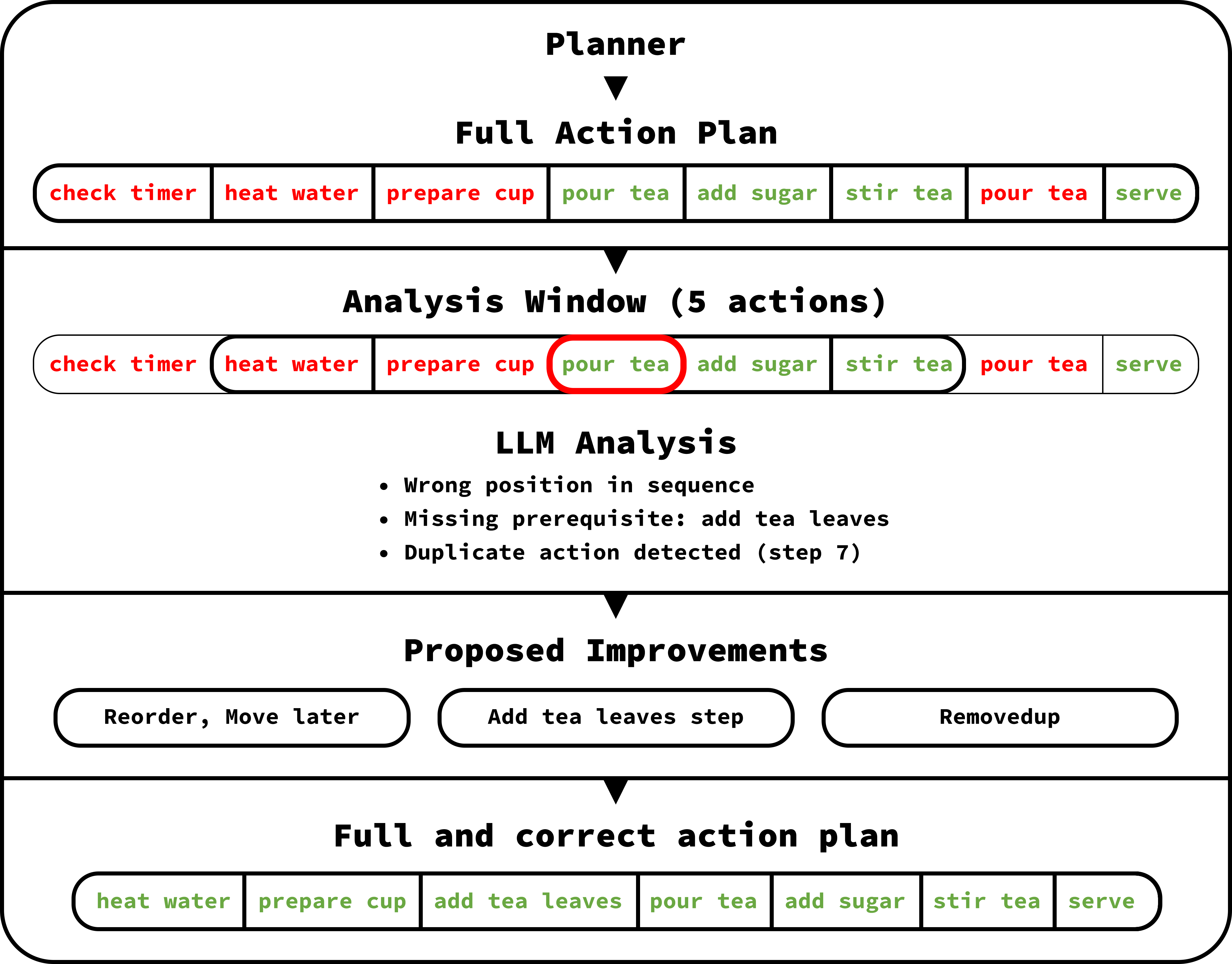}
\caption{VerifyLLM workflow for verification task plans. The system takes a generated action plan as input and analyzes it within the context window. Each window is processed by the LLM to identify position errors, missing prerequisites, and redundant actions. Based on this analysis, the system proposes improvements including reordering, adding necessary steps, and removing duplicates. The output is a refined action plan that maintains logical consistency and completeness. The example demonstrates verification of a tea preparation plan, where the system identifies and corrects issues with action ordering and missing prerequisites.}
\label{fig:visual_abstract}
\end{figure}
Researchers use Linear Temporal Logic (LTL) as a powerful formalism to specify robotic task requirements~\cite{kress2009temporal}. With its ability to express temporal relationships and logical constraints, LTL has been successfully applied to verify system properties in robotics~\cite{lahijanian2016iterative,aksaray2016distributed}, though these methods often face challenges with natural language understanding.
To address this challenge, we introduce VerifyLLM, a novel framework that integrates LTL with LLMs, as illustrated in Fig.~\ref{fig:visual_abstract}. Our system first translates action plans into LTL formulas, offering a formal representation that captures temporal dependencies and logical constraints.
The LLM then analyzes corresponding action sequences. This combination leverages the contextual understanding and common sense reasoning capabilities of LLMs~\cite{song2023llm} to detect potential errors before execution.
To rigorously assess our framework, we introduce two specialized datasets annotated with LTL specifications: ALFRED-LTL, VirtualHome-LTL. ALFRED-LTL is derived from the ALFRED dataset~\cite{ALFRED20}, which consists of household task instructions executed in a simulated environment. VirtualHome-LTL is adapted from VirtualHome~\cite{puig2018virtualhome}, a dataset containing human-like activities modeled in a virtual environment. 

Our main contributions include:
\begin{enumerate}
    \item The development of VerifyLLM, a novel verification framework that combines formal LTL representations with LLM-based reasoning to systematically identify inconsistencies in robot action plans. Our comprehensive evaluation demonstrates significant improvements in plan reliability across diverse household scenarios.
    \item The creation of two specialized datasets with LTL annotations: ALFRED-LTL, VirtualHome-LTL. 
\end{enumerate}

\section{RELATED WORKS}
\subsection{Plan Verification}
Traditional plan verification approaches have typically relied on model checking, theorem proving, and formal methods applied to PDDL representations. These classical techniques~\cite{cimatti2008informal,howey2004val} verify plans against domain specifications by systematically exploring state spaces to ensure all preconditions and effects are consistent. However, they often struggle with scaling to complex domains and lack the ability to incorporate common-sense knowledge that isn't explicitly encoded in domain definitions.
Recent research has increasingly explored the use of LLMs in robotic task planning. While most works focus on plan generation~\cite{huang2022language,ahn2022can}, relatively few address the critical challenge of plan verification.Current approaches primarily use LLMs to generate task plans from natural language instructions~\cite{huang2022language} or translate instructions into formal specifications~\cite{yang2024guiding}. Works like~\cite{zhang2023grounding} attempt to verify plans by detecting and recovering from execution failures, showing improved success rates on small-scale tasks. However, these methods rely on simple templates for describing failures (e.g., ``object X is blocking'') and struggle with complex scenarios involving temporal dependencies and safety constraints.Approaches such as~\cite{lin2023text2motion} and~\cite{yang2024guiding} focus on feasibility checking for generated plans but are limited to specific domains like motion planning. In ~\cite{li-etal-2024-llatrieval}, researchers introduced LLatrieval, a system where LLMs verify and iteratively improve retrieval results. Although focused on text generation, this verification approach parallels plan verification by using incremental checks to identify and correct errors.
In \cite{liang2024improving}, researchers proposed collaborative verification combining Chain-of-Thought and Program-of-Thought methods, demonstrating how different solution representations improve verification reliability. More specialized approaches have emerged, such as\cite{skreta2024replan}, which focuses on error recovery through continuous plan monitoring, and~\cite{grigorev2024common}, which addresses common-sense knowledge integration for robotics. While these methods address specific verification challenges, they tend to operate within limited contexts and don't fully integrate temporal reasoning, safety preconditions, and physical world constraints simultaneously. Our approach aims to bridge this gap through contextual analysis of plan segments.

The gap between plan generation and verification remains significant in robotics. Our VerifyLLM framework addresses this by systematically identifying three critical types of plan inconsistencies: position errors, missing prerequisites, and redundant actions. Unlike approaches in \cite{li-etal-2024-llatrieval} and \cite{liang2024improving}, we focus specifically on robotic task execution safety with actionable plan corrections that maintain commonsense consistency.

\subsection{Linear Temporal Logic}

Linear Temporal Logic (LTL)~\cite{pnueli1977temporal} has emerged as a powerful formalism for specifying robotic task requirements~\cite{kress2009temporal}. With its ability to express temporal relationships and logical constraints, LTL is well-suited for representing the structure and dependencies within complex task plans~\cite{pnueli1992temporal}. Traditionally, researchers apply LTL in model checking and theorem proving to verify system properties~\cite{lahijanian2016iterative,aksaray2016distributed}; however, these methods often face challenges when addressing the nuances of natural language understanding and real-world complexity. The seminal work in \cite{pnueli1977temporal} laid the foundation for using LTL in system verification, an approach later adapted to robotics in \cite{kress2009temporal}. Subsequent advancements, such as those in \cite{fainekos2009temporal}, introduced robust temporal logics that enable more realistic modeling of continuous systems.Recent research has begun to explore the integration of formal methods with learning-based approaches. For instance, \cite{desai2019soter} proposed frameworks for verifiable reinforcement learning that incorporate safety constraints, while \cite{grigorev2024common} introduced methods for common-sense verification through LLMs. These hybrid strategies aim to leverage the strengths of formal representations and natural language processing, albeit often within specific domains or simplified scenarios.In \cite{hsiung2022generalizing}, mapping natural language to lifted LTL representations was investigated to improve generalization across domains. This work highlights the potential of LTL as an intermediate representation that captures key temporal and logical relationships, facilitating transferability and deeper understanding in complex tasks. Similarly, in \cite{patel2020grounding}, grounding language to non-Markovian tasks using LTL was explored with a focus on learning rather than strict formal verification.

Overall, the research landscape indicates a need for approaches that combine the expressive power of LTL in representing temporal and logical dependencies with the language understanding capabilities of LLMs. While traditional formal methods emphasize strict verification, VerifyLLM  approach employs LTL as an intermediate representation to enhance the interpretability and refinement of generated task plans.

\section{PRELIMINARIES}
We use Linear Temporal Logic\cite{pnueli1977temporal} as our formal language for specifying robotic task requirements. LTL formulas follow this grammar:
\begin{equation}
\phi ::= p \mid \neg p \mid \phi_1 \land \phi_2 \mid \phi_1 \lor \phi_2 \mid \mathsf{G}(\phi) \mid \phi_1\mathsf{U}\phi_2 \mid \mathsf{F}(\phi)
\end{equation}
where $p \in P$ represents atomic propositions describing the robot's environment and possible states, $\phi$ denotes a task specification, and $\phi_1$, $\phi_2$ are LTL formulas. The basic logical operators include negation ($\neg$), conjunction ($\land$), and disjunction ($\lor$). The temporal operators serve distinct purposes: the globally operator $\mathsf{G}$ specifies that a property must hold at all future time points, the until operator $\mathsf{U}$ indicates that $\phi_1$ must hold continuously until $\phi_2$ becomes true, and the eventually operator $\mathsf{F}$ requires that $\phi$ must become true at some future time point.

\section{FORMAL TASK DEFINITION}

The classical planning problem is typically defined as a tuple
\begin{equation}
\mathcal{P} = \langle O, P, A, S, T, I, G, \tau \rangle,
\label{eq:planning-problem}
\end{equation}
where \(O\) denotes the set of objects in the environment, \(P\) represents the properties of these objects (e.g., their characteristics and affordances), \(A\) is the set of available actions, \(S\) is the set of possible states, \(T: S \times A \rightarrow S\) is the state transition function, \(I \in S\) is the initial state, \(G \subset S\) is the set of goal states, and \(\tau\) is the task description provided in natural language (see Eq.~\ref{eq:planning-problem} and Eq.~\ref{eq:plan-definition}).

Since our work focuses on the quality of generated plans, we introduce the concept of a \emph{plan} \(\pi\) as an ordered sequence of actions:
\begin{equation}
\pi = \langle a_1, a_2, \dots, a_n \rangle, \quad \text{with } a_i \in A,
\label{eq:plan-definition}
\end{equation}
and denote by \(\Pi\) the set of all admissible plans. For any given plan \(\pi\), the set of transitions(E is defined as
\begin{equation}
\{ (a_i, a_{i+1}) \mid 1 \leq i < n \}.
\label{eq:transitions}
\end{equation}

To ensure the correctness of transitions (Eq.~\ref{eq:transitions}) between actions with respect to the task \(\tau\), we extend the formulation to include a verification procedure. The extended planning problem is defined as
\begin{equation}
\mathcal{P} = \langle O, P, A, S, T, I, G, \tau, \Pi, V \rangle.
\label{eq:extended-planning-problem}
\end{equation}
The verification procedure \(V\) is designed to evaluate and refine the plan based on the problem description \(\tau\). Formally, the verification function is defined as
\begin{equation}
V: \Pi \times \tau \rightarrow \Pi,
\label{eq:verification-function}
\end{equation}
i.e., it takes as input an initial plan \(\pi_{\text{in}} \in \Pi\) and a task description \(\tau\), and returns a refined plan \(\pi_{\text{out}} \in \Pi\). The validity of the transitions is checked using common sense LLM reasoning combined with a formal specification \(\phi\) (see Eq.~\ref{eq:extended-planning-problem} and Eq.~\ref{eq:verification-function}).

\section{METHOD}

Plan verification for robots remains an underexplored challenge. While existing research focuses on plan generation, current verification methods have significant limitations: formal approaches struggle with natural language understanding and learning-based methods require extensive training data. To address these gaps, we propose VerifyLLM, a framework combining LLM with LTL for comprehensive pre-execution verification of robot task plans.

\begin{figure*}[t]
\centering
\includegraphics[width=\textwidth]{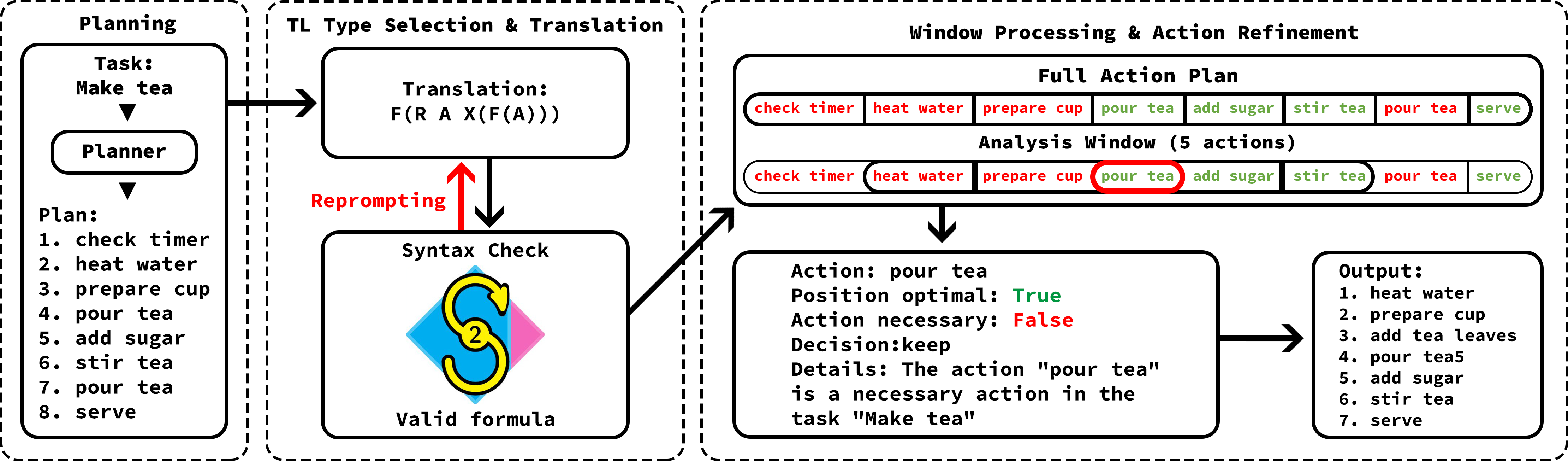}
\caption{The proposed LLM-based pre-execution validation framework. The system processes natural language plan through two main stages: (a) translation into formal temporal logic representations (LTL), (b) context common-sense enhancement.}
\label{fig:architecture}
\end{figure*}

\begin{algorithm}[t]
\caption{VerifyLLM Plan Verification}
\label{alg:verifyLLM}
\DontPrintSemicolon
\SetKwInput{KwInput}{Input}
\SetKwInput{KwOutput}{Output}
\SetAlgoLined
\LinesNumbered
\KwInput{Task description $\tau$, action sequence $\pi = \{a_1, ..., a_n\}$}
\KwOutput{Verificated sequence $\pi'$}

\SetKwFunction{FLTL}{TranslateToLTL}
\SetKwFunction{FVal}{verificateSequence}
\SetKwBlock{Begin}{begin}{end}

\Begin(\FLTL{$\tau$}){
    $E \gets$ translation examples\;
    \Return LLM($\tau$, $E$)\;
}

\Begin(\FVal{$\pi, \phi$}){
    $\pi' \gets \pi$\;
    \For{$i \gets 1$ \KwTo $|\pi'|$}{
        $\text{ctx} \gets \{\pi'[i-w:i+w]\}$ \tcp*{Extract window}
        $\text{props} \gets \text{ExtractProps}(\phi)$ \tcp*{Get propositions}
        
        \Switch{LLM(ctx, props)}{
            \Case{remove}{
                $\pi' \gets \pi' \setminus \{\pi'[i]\}$ \tcp*{Remove redundant action}
            }
            \Case{move}{
                $\pi' \gets \text{Reorder}(\pi', i)$ \tcp*{Fix position error}
            }
            \Case{augment}{
                $\pi' \gets \text{Insert}(\pi', i, \text{GenPrereq}(\pi'[i]))$ \tcp*{Add missing prerequisite}
            }
        }
    }
    \Return $\pi'$\;
}

$\phi \gets \FLTL(\tau)$ \tcp*{Convert task to LTL}
$\pi' \gets \FVal(\pi, \phi)$ \tcp*{Initial verification}
\While{not Converged($\pi'$)}{
    $\pi' \gets \FVal(\pi', \phi)$ 
}
\Return $\pi'$
\end{algorithm}

\subsection{VerifyLLM Architecture}

Our framework, illustrated in Figure~\ref{fig:architecture}, does not generate action plans but verifies pre-generated sequences. These plans are produced by an external planner, and our approach focuses on ensuring their logical correctness before execution. VerifyLLM enables systematic analysis of such task plans, addressing both logical consistency and contextual aspects before execution. The VerifyLLM architecture, presented in Algorithm~\ref{alg:verifyLLM}, comprises two main components: (1) a Translation Module that converts action plans into Linear Temporal Logic (LTL) formulas, and (2) a Verification Module that analyzes these plans using LLM-based reasoning enhanced by the LTL formalism.

The process begins with an action plan $\pi = \{a_1, a_2, ..., a_n\}$ and a natural language task description $\tau$. As shown in lines 1-4 of Algorithm~\ref{alg:verifyLLM}, the Translation Module first converts $\tau$ into an LTL formula $\phi$. This formula provides a formal representation of the task's temporal and logical constraints. The Verification Module (lines 5-23) then uses this formula to analyze the action plan through a sliding window approach, identifying and correcting three types of issues: incorrectly positioned actions, missing prerequisites, and redundant steps.
Consider the following example plan for making tea: check timer, heat water, prepare cup, pour tea, add sugar, stir tea, pour tea, serve.

Our VerifyLLM system would identify several issues with this plan:
\begin{itemize}
    \item \textbf{Redundancy}: Action 7 ``pour tea'' duplicates action 4, violating the non-redundancy principle
    \item \textbf{Missing prerequisite}: There's no action for adding a tea bag or tea leaves before pouring, which is a necessary prerequisite
    \item \textbf{Position error}: The preparation sequence is problematic—we need to add tea before pouring water
\end{itemize}

After verification, VerifyLLM would produce a corrected plan: check timer, heat water, prepare cup, add tea bag, pour hot water, add sugar, stir tea, serve.

This example demonstrates how VerifyLLM identifies and corrects logical inconsistencies in robot action plans by leveraging LLM reasoning capabilities guided by formal LTL constraints.

\subsection{Translation to LTL}

The Translation Module converts task descriptions into LTL formulas that capture temporal dependencies and logical constraints. This module uses a LLM with few-shot prompting to perform the translation.

As shown in lines 1-4 of Algorithm~\ref{alg:verifyLLM}, given a task description $\tau$, the module generates an LTL formula $\phi$ through the following process:
\begin{equation}
\phi = \text{LLM}(\tau, E) \text{ where } E = \{(e_1, \phi_1), ..., (e_k, \phi_k)\}
\label{eq:llm_translation}
\end{equation}

Here, $E$ represents a set of example pairs where each $e_i$ is a task description and $\phi_i$ is its corresponding LTL formula. The prompt structure is designed to guide the LLM in extracting key propositions and their temporal relationships from the task description (Eq.\ref{eq:llm_translation}).

For our simplified tea-making example ``Heat water, add tea, serve'' as shown in Fig.~\ref{fig:visual_abstract}, the module would generate formula (Eq.\ref{eq:ltl_example}):
\begin{equation}
\phi = \mathsf{F}(\text{heat\_water}) \land \mathsf{F}(\text{add\_tea}) \land \mathsf{F}(\text{serve})
\label{eq:ltl_example}
\end{equation}

We verify formula correctness using the \href{https://spot.lre.epita.fr}{Spot} library, which converts LTL formulas into Büchi automata for syntax validation. If errors are detected, the module employs a reprompting strategy to refine the formula up to three times.

\subsection{Verification Using LLM}

The verification Module, detailed in lines 5-23 of Algorithm~\ref{alg:verifyLLM}, analyzes action sequences using a sliding window approach. For a window size $w$ (typically set to 5 based on our empirical findings in Table~\ref{tab:window_results}), it examines consecutive action subsequences within the original plan $\pi$.

For each action $a_i$, the module examines its surrounding context:
\begin{equation}
\text{context}(a_i) = \{\text{prev}_{1...w}, a_i, \text{next}_{1...w}\}
\label{eq:context}
\end{equation}

where $\text{prev}_{1...w}$ and $\text{next}_{1...w}$ represent up to $w$ previous and upcoming actions within the window (lines 9-10 in Algorithm~\ref{alg:verifyLLM}). The context (Eq.\ref{eq:context}) also includes relevant atomic propositions extracted from the LTL formula $\phi$ (line 11 in Algorithm~\ref{alg:verifyLLM}). The example in Fig.~\ref{fig:prompt_image} demonstrates how the system analyzes action sequences for our tea-making scenario. The LLM identifies issues such as incorrect action ordering and redundant steps.This analysis evaluates four key aspects: position optimality (whether $a_i$ appears in the correct sequence), necessity (whether $a_i$ is essential for task completion) and compatibility (whether $a_i$ aligns with propositions from $\phi$).Based on this analysis, the module generates a verification decision (lines 12-21 in Algorithm~\ref{alg:verifyLLM}):
If the decision is to keep, the action remains unchanged. For move, remove, or augment decisions, the module performs the corresponding modification to the plan (lines 13-21 in Algorithm~\ref{alg:verifyLLM}).

\subsection{Prompt Engineering and Plan Optimization}

\begin{figure}[t]
\centering
\includegraphics[width=0.5\textwidth]{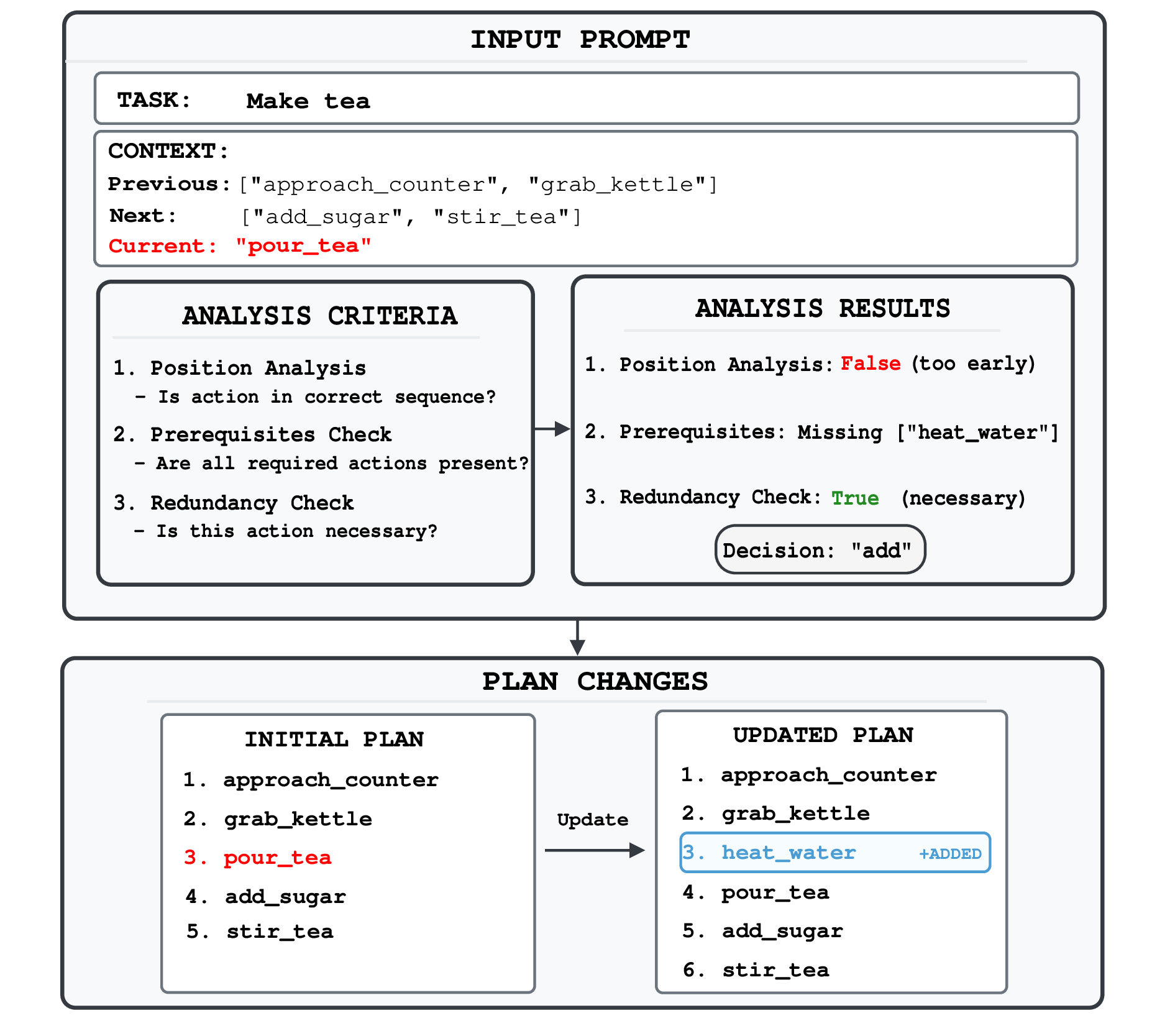}  
\caption{Prompt used for guiding the LLM-based reasoning in plan validation.}
\label{fig:prompt_image}
\end{figure}

Our verification process relies on structured prompts with three main components, as shown in Fig.~\ref{fig:prompt_image}. The input context provides task description, atomic propositions derived from LTL formulas, and the action window (previous actions, current action, and next actions within the sliding window). The analysis criteria guide the LLM through systematic verification of position correctness, prerequisite completeness, and action necessity.

In our tea example, the system would detect the missing prerequisite of adding tea before pouring water, identify the redundant second ``pour tea'' action, and suggest reordering steps into a logical sequence. The prompt enforces a structured JSON response containing verification decisions with detailed reasoning and suggested modifications, as illustrated in the ``Response Format'' component of Fig.~\ref{fig:prompt_image}.

Based on this analysis, the system employs three operations to correct invalid plans: removal of redundant actions (line 15 in Algorithm~\ref{alg:verifyLLM}) using $R(\pi) = \pi \setminus \{a_i \mid \exists j \neq i: \text{conflict}(a_i, a_j)\}$, addition of necessary preconditions (line 19) through $A(\pi, i, a_{\text{new}}) = \{a_1, \dots, a_i, a_{\text{new}}, a_{i+1}, \dots, a_n\}$, and reordering to optimize sequences (line 17).

Optimization continues iteratively until the plan end (lines 24-27 in Algorithm~\ref{alg:verifyLLM}). The system maintains interpretability by preserving the original intent while ensuring executability through modifications. 

\section{EXPERIMENTAL SETUP AND RESULTS}
Our experimental evaluation was designed to address several key aspects of the VerifyLLM framework: identifying common error patterns in LLM-generated plans, measuring the impact of our verification approach on plan quality, analyzing the contribution of individual components through ablation studies. For our experiments, we used the VirtualHome \cite{puig2018virtualhome} dataset from the ZeroShot planner \cite{huang2022language}. VirtualHome is a comprehensive benchmark that simulates household activities within a realistic virtual environment. It provides a diverse set of natural language task instructions along with corresponding action sequences for everyday tasks such as cooking, cleaning, and organizing. This dataset is richly annotated with details on object interactions and action dependencies, making it an ideal testbed for evaluating our plan verification and optimization methods. Prior to evaluation, we preprocess the action sequences by normalizing the actions (e.g., converting to lowercase and removing prepositions) to ensure consistent comparison across metrics. This comprehensive evaluation approach allows us to assess both the effectiveness of our approach and identify areas for improvement.

\subsubsection{Evaluation Metrics}
Our evaluation metrics for plan quality focus on overall sequence similarity and specific error types. One of the key metrics, \textbf{LCS Similarity (LCS)} metric quantifies the overall similarity between the original and optimized plan sequences. It is defined as:
\begin{equation}
\text{LCS} = \frac{|\text{LCS}(S_1, S_2)|}{\max(|S_1|, |S_2|)},
\label{eq:lcs_sim}
\end{equation}
where \(\text{LCS}(S_1, S_2)\) denotes the length of the longest common subsequence between the original plan \(S_1\) and the optimized plan \(S_2\). A value of 1 indicates identical sequences, while 0 indicates no commonality.

In addition, we introduce three error metrics for detailed analysis of plan quality. \textbf{Missing Actions} counts the number of required actions that are present in the reference plan but absent from the generated plan. Formally, if \(A_{\text{ref}}\) is the set of normalized actions from the reference plan and \(A_{\text{gen}}\) is the set from the generated plan, then the number of Missing Actions is given by:
\begin{equation}
\text{Missing Actions} = \left| A_{\text{ref}} \setminus A_{\text{gen}} \right|.
\label{eq:missing_actions}
\end{equation}

\textbf{Extra Actions} measures the number of unnecessary or redundant actions included in the generated plan that do not appear in the reference plan, defined as:
\begin{equation}
\text{Extra Actions} = \left| A_{\text{gen}} \setminus A_{\text{ref}} \right|.
\label{eq:extra_actions}
\end{equation}

\textbf{Order Errors} quantify discrepancies in the sequential ordering of actions. Let \(L_{\text{ref}} = [a_1^{\text{ref}}, a_2^{\text{ref}}, \dots, a_n^{\text{ref}}]\) and \(L_{\text{gen}} = [a_1^{\text{gen}}, a_2^{\text{gen}}, \dots, a_m^{\text{gen}}]\) be the ordered lists of normalized actions from the reference and generated plans, respectively. The number of Order Errors is then computed as:
\begin{equation}
\text{Order Errors} = \sum_{i=1}^{\min(n, m)} \mathbf{1}\{a_i^{\text{gen}} \neq a_i^{\text{ref}}\} + \max(0, m - n),
\label{eq:order_errors}
\end{equation}
where \(\mathbf{1}\{\cdot\}\) is the indicator function, equal to 1 if the condition is true and 0 otherwise.

\subsection{Results and Analysis}

\subsubsection{Common Types of Errors in Generated Plans}
To understand the typical failure modes in language model-generated plans, we analyzed 71 instructions from the work~\cite{huang2022language}. We evaluated plans generated by five different models ranging from 0.5B to 3B parameters. As shown in Table~\ref{tab:model_errors}, our analysis revealed consistent patterns of errors across all models regardless of their size.

\begin{table}[t]
\caption{Analysis of error types across different language models when generating task plans.}
\centering
\begin{tabular}{lcccc}
\toprule
Model & LCS & Missing & Extra & Order \\
\midrule
Llama 3.2 (3B) & 0.0875 & 10.28 & 10.45 & 17.58 \\
Llama 3.2 (1B) & 0.0717 & 10.28 & 9.14 & 16.48 \\
Qwen 1.5 (0.5B) & 0.0690 & 10.21 & 9.01 & 16.73 \\
Qwen 2.5 (3B) & 0.0680 & 10.29 & 7.51 & 14.74 \\
Kanana Nano (2.1B) & 0.0650 & 10.28 & 8.80 & 13.94 \\
\bottomrule
\end{tabular}
\label{tab:model_errors}
\end{table}

Our findings identified three consistent types of plan errors across all tested models. \textbf{Missing Actions} (Eq.~\ref{eq:missing_actions}) were the first major issue: models consistently omitted 10-11 necessary actions per plan on average, including safety checks, preparatory steps, and environmental preconditions that humans would naturally consider. The second type was \textbf{Extra Actions} (Eq.~\ref{eq:extra_actions}), where plans contained 7-10 superfluous actions on average that either added unnecessary complexity or could potentially interfere with successful task completion. \textbf{Order Errors} (Eq.~\ref{eq:order_errors}) emerged as the most prevalent issue, with 14-18 ordering mistakes per plan on average, indicating a significant challenge in understanding temporal dependencies. The Longest Common Subsequence (LCS) score, all below 0.09, further demonstrates the substantial divergence from optimal plans. These findings directly motivated the design of our verification system, particularly the inclusion of mechanisms for detecting missing prerequisites, identifying redundant actions, and optimizing action ordering.

\subsubsection{Impact of Verification Module}

We used Llama-3.2-1B for verification across different types of baselines to evaluate the effectiveness of our approach. This allowed us to systematically assess how different verification methods perform when implemented with the same underlying language model.

We compared our VerifyLLM approach with three alternative verification methods to establish a comprehensive baseline:

The \textbf{Baseline Optimizer} examines adjacent action pairs and suggests insertions based on direct analysis of consecutive steps, using a straightforward prompt structure. The \textbf{Chain of Thought (CoT) Optimizer} employs explicit step-by-step reasoning to analyze transitions between actions, prompting the LLM to consider action goals, identify logical gaps, and reason about prerequisites. The \textbf{Window-based Optimizer} extends CoT by incorporating contextual information, analyzing actions within a sliding window to consider broader context when suggesting modifications.

Table~\ref{tab:plan_comparison} presents a comprehensive comparison of these approaches alongside our VerifyLLM method using both Llama-3.2-1B and Claude as underlying models.

\begin{table}[t]
\caption{Comprehensive comparison of different verification approaches}
\centering
\begin{tabular}{lcccc}
\toprule
Method & LCS & Missing & Extra & Order \\
\midrule
Baseline Opt. & 0.0656 & 10.38 & 11.40 & 19.04 \\
CoT Opt. & 0.0705 & 10.38 & 9.35 & 16.80 \\
Window Opt. & 0.0623 & 10.38 & 13.47 & 22.84 \\
VerifyLLM (Llama) & 0.0982 & 11.18 & 9.13 & 15.12 \\
VerifyLLM (Claude) & \textbf{0.183} & \textbf{10.17} & \textbf{8.32} & \textbf{9.47} \\
\bottomrule
\end{tabular}
\label{tab:plan_comparison}
\end{table}

The results reveal important insights about different verification approaches. Overall, the three baseline methods showed limited effectiveness in improving plan quality. While they reduced the percentage of missing actions compared to the original Llama-3.2-1B results, they degraded performance across other key metrics. Among the baseline methods, CoT Optimization achieved the highest LCS score (0.0705), but this still represents poor sequence similarity to ground truth. All three baseline methods maintained identical missing action scores (10.38), suggesting a fundamental limitation in identifying and addressing missing prerequisites.

The baseline approaches performed particularly poorly in terms of extra actions and ordering errors. Window Optimization significantly worsened the plan structure by introducing excessive extra actions (13.47) and substantially increasing ordering errors (22.84) compared to the original plan generation. Even the best-performing baseline, CoT Optimization, still introduced more unnecessary actions and ordering errors than would be acceptable for reliable robotic task execution.

In contrast, our full VerifyLLM approach, especially when implemented with Claude, demonstrated substantial improvements across all metrics. The Claude-based implementation achieved a much higher LCS similarity (0.183), reduced missing actions (10.17), significantly decreased extra actions (8.32), and most notably, reduced ordering errors by nearly 40\% (9.47) compared to the best baseline method. These results clearly demonstrate that combining LLMs with formal logical guidance through LTL significantly enhances verification quality.

\subsubsection{Ablation Studies}
To understand the contribution of each component in our system, we conducted comprehensive ablation studies. We systematically removed or modified key components and evaluated the impact on system performance. Table~\ref{tab:ablation_results} presents the results of our ablation experiments.

\begin{table}[t]
\caption{Ablation study results showing the impact of different components.}
\centering
\begin{tabular}{lcccc}
\toprule
Configuration & LCS Sim. & Missing & Extra & Order \\
\midrule
Full System & \textbf{0.183} & \textbf{10.17} & 8.32 & \textbf{9.47} \\
No LTL Translation & 0.178 & 10.25 & \textbf{8.25} & 9.80 \\
No LLM Verification & 0.0717 & 10.28 & 9.14 & 16.48 \\
\bottomrule
\end{tabular}
\label{tab:ablation_results}
\end{table}

The ablation studies provided several important insights. Removing the \textbf{LTL translation module} resulted in only a modest degradation in performance: LCS similarity decreased slightly from 0.183 to 0.178, and ordering errors increased from 9.47 to 9.80. This indicates that while the LTL module contributes to capturing temporal relationships between actions, its overall impact is relatively limited. In contrast, eliminating the \textbf{LLM-based verification component} had a dramatic effect on plan quality. Specifically, its removal led to a 74\% increase in ordering errors (from 9.47 to 16.48) and a 60\% decrease in LCS similarity (from 0.183 to 0.0717). These findings confirm that both formal logical guidance provided by the LTL module and the robust reasoning capabilities of the LLM are critical for effective plan verification. In our full system, the combination of these components offers complementary strengths that address different aspects of the verification process.

\subsubsection{Window Size Analysis}
We experimented with different sliding window sizes to determine the optimal context length for processing plans. Table~\ref{tab:window_results} presents the F1-scores achieved with window sizes of 3, 5, and 7 actions on the Zero-Shot dataset.

\begin{table}[t]
\caption{Module performance across different window sizes.}
\centering
\begin{tabular}{lcc}
\toprule
Dataset & Window Size & F1-score \\
\midrule
Zero-Shot & 3 & 0.58 \\
Zero-Shot & 5 & \textbf{0.65} \\
Zero-Shot & 7 & 0.54 \\
\bottomrule
\end{tabular}
\label{tab:window_results}
\end{table}

The results highlight that a window size of 5 yields the best F1-score (0.65), outperforming both smaller (3) and larger (7) windows. This suggests that an intermediate window size provides an optimal balance, capturing sufficient contextual information while avoiding excessive complexity or dilution of relevant details. A window that is too small (size 3) lacks adequate context for complex dependencies, while a larger window (size 7) introduces noise and irrelevant information that can confuse the model. Based on these findings, we selected a window size of 5 for all subsequent experiments.

\subsubsection{Motivation for Transitioning to Larger Models}
Our experimental results strongly motivate a transition from smaller to larger language models for plan verification. Smaller models struggle with the intricate, nested vocabulary Fig.~\ref{fig:prompt_image} required for comprehensive plan analysis. They often fail to capture long-range dependencies and nuanced contextual relationships, leading to suboptimal handling of multi-layered prompts. In contrast, larger models (e.g., those with 3B parameters or the Claude variant) demonstrate enhanced capabilities in understanding complex temporal dependencies and logical relationships.

As shown in Table~\ref{tab:plan_comparison}, the VerifyLLM system implemented with larger models achieves significantly higher LCS similarity scores and lower ordering errors, resulting in plans that more closely resemble the optimal structure. This performance gain is attributable to the larger models' superior ability to process and understand multi-level nested prompts, making them better suited for tasks requiring detailed and hierarchical plan analysis.

\section{CONCLUSION AND FUTURE WORK}
VerifyLLM combines Large Language Models with verification methods to address three critical plan inconsistencies: position errors, missing prerequisites, and redundant actions. Our key contributions include a framework for translating instructions into Linear Temporal Logic formulas, a contextual analysis approach for validating action sequences, specialized datasets with LTL annotations. Experimental results demonstrate significant improvements in plan quality and verification accuracy across household domains, with consistent error patterns observed across different language models. Despite promising results, we acknowledge limitations including challenges with complex temporal dependencies and parallel task sequences. The system also needs better mechanisms for mapping between robot capabilities and generated plans. Future work will focus on handling complex temporal relationships, improving parallel task processing, and expanding beyond household domains to industrial, healthcare, and outdoor environments.

\bibliographystyle{IEEEtran}
\bibliography{references}
\end{document}